\documentclass{article}

\usepackage{PRIMEarxiv}

\usepackage[utf8]{inputenc} 
\usepackage[T1]{fontenc}    
\usepackage{hyperref}       
\usepackage{url}            
\usepackage{booktabs}       
\usepackage{amsfonts}       
\usepackage{nicefrac}       
\usepackage{microtype}      
\usepackage{lipsum}
\usepackage{fancyhdr}       
\usepackage{graphicx}       
\graphicspath{{media/}}     
\usepackage{multirow}
\usepackage{amsmath,amssymb,amsfonts}
\usepackage{amsthm}
\usepackage{mathrsfs}
\usepackage[title]{appendix}
\usepackage{xcolor}
\usepackage{textcomp}
\usepackage{manyfoot}
\usepackage{booktabs}
\usepackage{listings}
\usepackage{subfig}
\usepackage{tabularx}
\usepackage{algorithm}
\usepackage{algorithm}
\usepackage{algpseudocode}
\usepackage{colortbl}

\definecolor{lightblue}{RGB}{173,216,230}
\DeclareMathOperator*{\argmin}{\arg\!\min}%

\title{Condensed-Gradient Boosting}

\author{
  Seyedsaman Emami \\
  Escuela Polit\'ecnica Superior \\
  Universidad Aut\'onoma de Madrid \\
  Madrid\\
  \texttt{emami.seyedsaman@uam.es} \\
   \And
  Gonzalo Mart\'{\i}nez-Mu\~noz \\
  Escuela Polit\'ecnica Superior \\
  Universidad Aut\'onoma de Madrid \\
  Madrid\\
  \texttt{gonzalo.martinez@uam.es} \\
}

\begin{document}
\maketitle

\begin{abstract}
This paper presents a computationally efficient variant of Gradient Boosting
    (GB) for multi-class classification and multi-output regression tasks.
    Standard GB uses a 1-vs-all strategy for classification tasks with more than
    two classes. This strategy entails that one tree per class and iteration has
    to be trained. In this work, we propose the use of multi-output regressors as
    base models to handle the multi-class problem as a single task. In addition,
    the proposed modification allows the model to learn multi-output regression
    problems. An extensive comparison with other multi-output based Gradient
    Boosting methods is carried out in terms of generalization and computational
    efficiency. The proposed method showed the best trade-off between
    generalization ability and training and prediction speeds.
    Furthermore, an analysis of space and time complexity was undertaken.
\end{abstract}


\section{Introduction}

The field of machine learning has rapidly advanced in recent years, with
numerous applications in a variety of industries. In particular, ensemble
learning models have proven to be effective in solving multi-class
classification \cite{Emami2023icip,Asif2023}, multi-output regression
problems \cite{Kucuk2022,Emami2022esann,Emami2024}, Multi-objective problems
\cite{yu2018multiobjective}, and multi-task learning
\cite{Emami2023hais,Jeong2023}. Ensemble learning techniques consistently
demonstrate exceptional robustness and accuracy
\cite{Kun2022,Demir2023,Namamula2024}, ranking them among the
top-performing
methods \cite{comparison2017zhang,bentejac21comp}.
Moreover, ensemble models exhibit superior performance in contrast to deep
learning frameworks within model training and evaluation concerning tabular
datasets, necessitating lesser tuning efforts \cite{SHWARTZZIV2022}. Ensemble
models leverage the combination of multiple base learners, typically weak
individually, to create a robust learner. This fusion generally improves the
generalization ability
of its members.

One of the most renowned ensemble models is Gradient Boosting (GB), often
referred to as Gradient Boosted Decision Trees (GBDT) \cite{friedman2001greedy}.
In GB, the learning method works by sequentially creating new base models that
learn the portion of the concept not 'captured' by previous learners. The
problem is posed as an optimization-problem in which the output of each new
model is built to be correlated with the negative gradient of the loss function
of previous iterations. The current implementation of GB has both advantages and
disadvantages, and significant improvements in this area have been achieved in
recent years. For instance, in \cite{chen2016xgboost}, a fast variant of GB was
presented called XGBoost.
XGBoost integrates a new regularization method into the objective function of
gradient boosting and employs a fast strategy for decision tree splitting.
This method drew much attention for achieving the best performance in many
Kaggle competitions \cite{chen2016xgboost}. In a separate study, CatBoost
tackles the prediction shift phenomenon observed during gradient boosting
training by introducing a shifted permutation technique
\cite{prokhorenkova2017catboost}. Moreover, CatBoost optimizes the training
process, focusing particularly on categorical features. Furthermore, LightGBM
\cite{ke2017lightgbm} significantly advances Gradient Boosting techniques. It
introduces two innovative methodologies: Gradient-based One-Side Sampling (GOSS)
and Exclusive Feature Bundling (EFB). These methodologies are designed to
address the inherent challenges in training speed encountered by earlier
gradient boosting models, especially when dealing with large-scale datasets.

One limitation of the current gradient boosting approaches is that, for
multi-class classification, it requires training one classifier for each class
per iteration. That is, the model optimizes each class in a separate process.
This approach can lead to substantial computational overhead, particularly when
dealing with a large number of classes. Furthermore, it often results in complex
and slower ensembles. Moreover, for multi-output regression problems, it is
required to build one isolated ensemble for each output, ignoring the
correlations between the dependent variables. As the correlations between
multiple outputs exist, \cite{zhang2012multi} demonstrates that these
correlations can enhance the performance of multi-output Support Vector
Regression. Moreover, the complexity of existing ensemble models, particularly
in the context of multi-class classification and multi-output regression, is a
noteworthy concern. The conventional practice of training a separate classifier
for each class in multi-class problems can lead to a proliferation of model
components, significantly impacting computational demands. These large
ensembles prolong both the training phase and prediction times
\cite{PeterDiego2017,Lu2020}.
Furthermore, certain studies have proposed strategies to manage the
computational demands of ensemble complexity. For instance, employing techniques
like dimensionality reduction can expedite achieving the desired accuracy, as
demonstrated by \cite{ZORARPACI2024}. Alternatively, increasing computational
resources may also be necessary. Hence, the imperative arises to develop less
time complex and more fast ensembles \cite{Huang2024}.
Moreover, the extensive number of individual models can make both model
interpretability and visualization more challenging. In the case of multi-output
regression, the requirement to construct distinct ensembles for each output
variable further exacerbates the issue. Each ensemble adds to the overall
complexity, rendering the process of model management and optimization a
challenging task.
In addition to the interest in developing multi-output regression models
capable of effectively incorporating the correlation between output targets
in addition to input features
\cite{zhang2012multi,Borchani2015},
it has been observed that ensemble variants of multi-output models exhibit
very good performance across a large range of multi-output regression
datasets, spanning disciplines from aerospace engineering and energy fields to
environmental
studies \cite{spyromitros2016multi, Spyromitros20200, Emami2024}.
To address these challenges, it
becomes imperative to explore strategies that streamline the training and
prediction processes, while maintaining, or even enhancing, the overall model
performance. The development and evaluation of Condensed-Gradient Boosting
(C-GB), presented in this study, represents a step toward mitigating the
complexities associated with current models and achieving more efficient
solutions in multi-class classification and multi-output regression scenarios.

There are some proposals to use a single multi-output tree for all classes at
each boosting iteration to produce faster ensembles. In cited works by
\cite{ponomareva2017compact,zhang2020gbdt}, the the proposed methodologies are
developed within the framework of gradient boosting, specifically leveraging the
XGBoost algorithm and LightGBM. The two referenced studies employ the
second-order Taylor expansion method to approximate their respective objective
functions. Notably, the derivations are adapted to accommodate vectorized
expressions of this expansion.

The objective of the current study has two parts. Firstly, it introduces a new
GB model that addresses various problems uniformly, including binary/multi-class
classification and single/multi-output regression. This is accomplished by
employing a single model, potentially with multiple outputs, at each step.
In contrast to \cite{ponomareva2017compact,zhang2020gbdt}, the optimization of
the proposed method is based on a two-step procedure. First, we fit the base
learner by least-squares to the pseudo-residuals and then approximate the
solution with the Newton-Raphson step following the original GB strategy
\cite{friedman2001greedy}. The second objective of this paper is to perform an
exhaustive comparison of multi-output GB methods in terms of generalization
accuracy and computational efficiency.

The present paper's outline is as follows: Section \ref{sec:related_work}
overviews relevant prior research. Section \ref{sec:methodology} explores the
construction of Gradient Boosting, detailing the formulated objective function
and algorithm designed for tackling multi-class classification and multi-output
regression problems. Additionally, it provides illustrative examples to clarify
the operational principles of the proposed model. In Section
\ref{sec:experiments}, the different subsections outline: the dataset utilized
for the experiments, the code repositories corresponding to the studied models,
the experimental setup, the evaluation metrics employed, the experimental
methodology, and the results of various empirical analyses. Lastly, Section
\ref{sec:conclusion} summarizes the conclusions of the paper.

\section{Related work}
\label{sec:related_work}

Several studies have proposed different extensions
to boosting-like methods to
better handle multi-class classification.
These extensions are based on one of
the two main families of boosting methods:
AdaBoost \cite{freund1997decision,
    mason1998direct, mason1999functional} and
Gradient Boosting (GB)
\cite{friedman2001greedy, friedman2003multiple,
    chen2016xgboost}.

The original AdaBoost \cite{freund1997decision}
algorithm proposes two boosting
variants to handle multi-class
classification (called AdaBoost.M1 and M2).
AdaBoost.M1 is based on using multi-class
weak base learners, while AdaBoost.M2
converts the multi-class problem into
a set of binary tasks.
These multi-class versions could
have theoretical and practical convergence
problems \cite{freund1997decision,eibl2005multiclass}.
The work proposed in
\cite{eibl2005multiclass} tackles these problems by
defining an exponential bound on a test
measure called pseudo-loss error, in
which each base learner (decision
stumps in their work) minimizes its upper bound
with gradient descent.
Their method proved better convergence than
AdaBoost.M2.

In multi-class tasks,
the approach followed by Gradient Boosting
\cite{friedman2000additive} is to optimize a
multi-class loss function separately for
each class by training one tree per class.
The main drawback of this approach is
the complexity of the solution, as
it involves generating a set
of trees at each iteration.

In \cite{li2009abc}, an adaptive base class
boosting (ABC-MART) is presented.
The proposal derives the multi-class
logistic regression loss function
\cite{friedman2001greedy} given a sum-to-zero
constraint, such that the sum of the
outputs of the models adds up to zero.
This allows them to choose, at each epoch,
the base class which obtains the highest reduction
of the training loss in a
greedy manner. However, this greedy
approach introduces higher training
complexity in the algorithm.

The most similar studies to the present work
are \cite{ponomareva2017compact,zhang2020gbdt}.
The idea of those studies is the
same as that of this work, which is to train
a single tree for all classes at
each boosting iteration, although they
are based on a different optimization
procedure.
The work of \cite{ponomareva2017compact} is based
on XGBoost \cite{chen2016xgboost}.
Their implementation, introduced as
open-source based
on TensorFlow, is named TFBT
(stands for TensorFlow Boosted Trees) model.
They showed how the use of a single multi-output
regressor at each step produces less complex
ensembles, reducing the size by approximately a
factor of the number of classes.
They introduced two strategies in terms of loss
function calculation. One of them considers
the full Hessian loss, and the other
assumes that the Hessian matrix is
diagonal, leading to a faster algorithm.
Additionally, they included a layer-by-layer
boosting methodology, characterized by
incremental adjustments within the function space.
In a similar approach, \cite{zhang2020gbdt}
introduced a Gradient-Boosted
Decision Trees for Multiple Outputs
(GBDT-MO) by training a single tree per
iteration. The use of the objective
function and its derivatives is the same as
in \cite{ponomareva2017compact}.
Although GBDT-MO is implemented on top of
LightGBM \cite{ke2017lightgbm} and XGBoost
objective function \cite{chen2016xgboost}, they
extended the histogram approximation of
LightGBM for the multiple outputs,
they also adapted the algorithm to
multi-output regression and
multi-class classification.
These two studies
(\cite{ponomareva2017compact,zhang2020gbdt})
consider the vector
form of Taylor expansion to approximate their
objective function, following the
optimization procedure of XGBoost.
Note that the original XGBoost
and LightGBM also handle multi-output problems
but build one tree per output
per iteration, while the studies by
\cite{zhang2020gbdt} and
\cite{ponomareva2017compact} use a single tree per
iteration.
In contrast,
our proposed model is
a two-step optimization procedure that applies
gradient descent and a Newton-Raphson step
approximation to find the optimal solution,
following the original GB approach. In the current
study, we demonstrate that the proposed approach
enhances generalization ability, training
and prediction speeds, and model complexity.

A recent application of multi-output
regression using gradient boosting involves
predicting multi-step-ahead traffic
speeds simultaneously \cite{zhan2020multi}.
Three distinct approaches were employed to
address the multi-output regression in \cite{zhan2020multi}.
The first is an iterative strategy that uses predicted
values for subsequent output predictions.
The second approach is a direct strategy that
constructs independent models for each
output ignoring inter-output correlations.
Lastly, the last strategy involved establishing
a multivariable correlation matrix for prediction variables to
compute correlation coefficients among sequences for
each prediction step. This approach is then
used to update the impurity function and consider
all the outputs in the split finding.

Moreover, there are several implementations of
multi-output decision trees,
such as \cite{blockeel2000top,struyf2005constraint}.
The work presented in \cite{blockeel2000top}
is designed in the context of clustering. It uses
a generic prototype distance function to
maximize the inter-cluster distance in
a standard top-down decision tree
approach \cite{breiman1984classification}.
The functionality of their decision
tree varies based on the selected prototype
function, enabling its application in
both classification and regression tasks.
In \cite{struyf2005constraint}, a constrained multi-output decision tree
model is presented within the framework initially proposed.
Their method entails the training of an extensive multi-output decision tree,
followed by a pruning phase aimed at fulfilling specified user constraints
pertaining to both size and accuracy.
These multi-output decision trees can be combined into ensembles
\cite{kocev2007ensembles}. In \cite{kocev2007ensembles}, the multi-output decision trees of
\cite{struyf2005constraint} are combined with bagging \cite{breiman1996bagging}
and random forests \cite{breiman2001random}. The proposed ensembles
\cite{kocev2007ensembles} are tested on various classification and regression
tasks, showing better performance than the single base models. Moreover, in the
experiments carried out, the proposed multi-objective ensemble performed better
on regression tasks than the single-objective ensembles. In classification, the
performance of the multiple and single-output models was equivalent.

In a recent study by \cite{NAKANO2022} a deep tree-ensemble (DTE) model is
proposed. DTE approach entails constructing an ensemble of multiple layers of
random forests (referred to as deep forest), with each layer enriching the
initial feature set through the incorporation of a representation learning
component based on tree embeddings. The DTE model is specifically developed for
multi-output regression and multi-class classification tasks. Our proposed model
diverges from the approach presented in \cite{NAKANO2022}. We utilize
multi-output decision tree regressors trained in an additive manner.
In \cite{Emami2022esann}, a novel multi-output regression training procedure for
shallow neural networks based on gradient boosting is proposed.
This procedure demonstrates promising performance in terms of accuracy for
multi-output regression tasks.
Moreover, that study is extended in \cite{Emami2024} proposing an
ensemble of multi-output regression deep neural networks.
The experimental results show that the proposed deep model achieves better
generalization accuracy across various multi-output regression problems with
respect to the shallow models.
In the current paper, a multi-output decision tree regressor is introduced as the
additive model within gradient boosting, instead of using hidden unit of a
shallow neural network. This approach produces simpler and faster to train
models.

\section{Methodology}
\label{sec:methodology}

This section discusses the
Gradient Boosting (GB) algorithm,
considering a dataset $D=\{(\mathbf{x}_i,
    y_i)\}_{i=1}^N$ drawn from a distribution
$\mathcal{D}=\{ \mathcal{X}, \mathcal{Y} \} $
where $\mathcal{X} \in \mathbb{R}^P$ and
$\mathcal{Y} \in \mathbb{R}^K$ refer to a $P$-dimensional
feature space
and a $K$-dimensional output space, respectively.
The algorithm is reviewed based on
the Greedy function approximation
\cite{friedman2001greedy} (Subsection \ref{subsec:gb}).
Furthermore, proposed modifications to
handle multi-class classification by
considering all classes simultaneously
at each step, as well as outputs for
multi-output regression are described (Subsection
\ref{subsec:cgb}).
In this study, $M$ represents the
total count of boosting epochs,
with $m$ serving as the index variable
for individual boosting epochs within
each iteration.

\subsection{Gradient Boosted Decision Trees}
\label{subsec:gb}

In order to build the functional relationship
between $\mathbf{x}_i$ and $y_i$ as
$\hat{F}(x)$, that minimizes a given loss function ($L$)

\begin{align}
    \hat{F}(\mathbf{x}) & =
    \argmin_{F(\mathbf{x})}
    \mathbb{E}_{\mathbf{x}, y}\big[L(y, F(\mathbf{x}))\big],
\end{align}
where $\mathbb{E}$ is the expected value over the joint distribution,
the optimization procedure followed by GB
consists of estimating the objective function
through
an additive expansion
\begin{align*}
    \hat{F}(\mathbf{x})  =
    \sum_{m=0}^{M}
    \gamma_m h_m(\mathbf{x}),
\end{align*}
where $h(\mathbf{x})$ is a base regressor
model. The model, $\hat{F}(\mathbf{x})$,
is created in a stagewise greedy process as
\begin{equation}
    \label{eq:mm}
    (\gamma_m, h_m) =
    \argmin_{\gamma, h} \sum_{i=1}^{N}
    L\big(y_i, F_{m-1}(\mathbf{x}_i) +
    \gamma h_m(\mathbf{x}_i)\big),
\end{equation}
and
\begin{equation}
    \label{eq:additive}
    F_m(\mathbf{x}) = F_{m-1}(\mathbf{x}) +
    \nu \gamma_m h_m(\mathbf{x}),
\end{equation}
where $\nu$ is the learning rate hyperparameter
that is useful for regularizing the model.
In order to optimize Eq. \eqref{eq:mm}, first the model $h$
is trained, and then the
value of $\gamma$ is optimized.
At each step, a gradient descent step is
performed by computing the negative gradient, or pseudo-residuals, for each
data point
\begin{equation}
    \label{eq:pseudo}
    r_{\{i, m\}} = - \Bigg[\frac{\partial L(y_{i},
            F(\mathbf{x}_i))}{\partial F(\mathbf{x}_i)}\Bigg]_
    {F(\mathbf{x}) = F_{m-1}(\mathbf{x})}.
\end{equation}

The $m$-th decision tree regressor model,
$h_m(\mathbf{x})$, is
trained to adjust to the pseudo-residuals or gradients.
Note that the gradients are optimized using
a constrained function $h$,
usually a decision tree, that is built using the
square loss
regardless
of
the loss function we are trying to optimize
\begin{equation}
    h_m = \argmin_{h} \sum_{i=1}^{N}
    \big(r_{im} - h(\mathbf{x}_i)\big)^2.
\end{equation}
Hence, $\gamma$ is subsequently optimized on
the original loss
function

\begin{equation}
    \label{eq:tree}
    \gamma_m = \argmin_{\gamma}
    \sum_{i=1}^{N}L\big(y_i,
    F_{m-1}(\mathbf{x}_i) + \\
    \gamma h_m(\mathbf{x}_i)\big).
\end{equation}

The approximation is performed
using a Newton-Raphson step.
When the model $h$ is a decision tree,
this line search
optimization can be done independently for each terminal
leaf,
because a decision tree can be seen
as an additive model of the outputs of the
terminal regions:
$h_m(\mathbf{x}_i) = \sum_{j=1}^J b_{mj} \mathbb{I}(\mathbf{x}_i \in
    \mathbb{R}_j)$, where $J$ is the number of terminal nodes of the tree,
$\mathbb{R}_j$ is the region covered by leaf $j$ and $b_j$ is its output. The
terminal regions are disjoint and cover the whole data space.
Consequently, a different $\gamma$ value can be computed for each terminal
node $j$. We can integrate the general $\gamma$ constant in each node
as $\gamma_{\{m, j\}}=\gamma_m b_{\{m, j\}}$. With this, in the cases of binary
classification or single-output regression in which one decision tree is built
at each $m$ iteration, the output of the leaf nodes is updated as

\begin{equation}\label{eq:TerminalRegions}
    \gamma_{\{m, j\}} =  \argmin_{\gamma} \sum_{\mathbf{x}_i \in \mathbb{R}_{\{m, j\}}} L\big(y_{i},
    F_{m-1}(\mathbf{x}_i) + \gamma \big),
\end{equation}
Note that for regression problems in which the square loss is optimized, this
last step is not necessary as the leaf nodes already have the correct output.

Recent versions of Gradient Boosting include randomization
hyperparameters that help its generalization performance and serve as a
regularization mechanism \cite{Jerome2002}. These randomization techniques
include data subsampling and feature subsampling for split finding.

\subsection{Condensed Gradient Boosting}
\label{subsec:cgb}

For the case of multi-class classification,
since the base decision tree regressors supply scalar values, gradient boosting
ensemble needs to build one tree per class and iteration.
In this
subsection, we derive the extension needed for GB \cite{friedman2001greedy} to
work with a single decision tree regressor with vector-size outputs for multi-class
classification and multi-output regression. One of these trees could replace
the single-output trees used in standard GB.
Considering a dataset as $D=\{(\mathbf{x}_i, y_i)\}_{i=1}^N$ and
selecting a multi-class label with $K$
class labels $k \in [1, K]$ sampled
from the $\mathcal{D}$ distribution,
a one-hot encoding approach is employed to
transform the multi-class labels into a logical matrix.
In this matrix, each row $i$ represents a
vector of $y_{\{i, k\}}$, where $y_{\{i, k\}}$
equals $1$ if the $i$-th instance belongs to class $k$,
and $0$ otherwise.
The objective is to find $K$ functions
$\mathbf{F}=\{F_k\}_{k=1}^K$
that minimize a given loss
function in an additive manner using multi-output models
\begin{equation}
    \label{eq:addmulti}
    \mathbf{F}_m(\mathbf{x}) = \mathbf{F}_{m-1}(\mathbf{x}) +
    \nu \tilde{\mathbf{h}}_m(\mathbf{x})
\end{equation}
where $\tilde{\mathbf{h}}_m(\mathbf{x}) = \{\gamma_{\{k, m\}}
    h_{\{k, m\}}(\mathbf{x})\}_{k=1}^K$ is the multi-output model with $K$ outputs, and
$\nu$ is the learning rate hyperparameter.
Using the one-hot encoded
class labels,
at each iteration, a single multi-output tree is trained
to fit the data
in which the values to learn are
a matrix of size $N \times K$ with the residuals of the previous predictions
\begin{equation}\label{eq:pseudo_cgb}
    r_{\{i, k, m\}} = - \Bigg[\frac{\partial L(y_{\{i,k\}},
            F_k(\mathbf{x}_i))}{\partial F_k(\mathbf{x}_i)}\Bigg]_{F_k(\mathbf{x}) =
        F_{\{k,m-1\}}(\mathbf{x})}.
\end{equation}
For node $j$ with a vector of $K$ outputs of the decision tree regressor, the
criterion to minimize forthcoming splits is the average of Mean Squared Error
(MSE) over the $K$ outputs,
\begin{equation}
    \label{eq:quality}
    Q(j) = \frac{1}{N_j K} \sum_{k=1}^{K} \sum_{\mathbf{x}_i \in \mathbb{R}_j}^{}(y_{ik} - \bar{y}_{k})^2 ,
\end{equation}
where $\bar{y}_k$ is the mean value of target $k$ in the region $j$, and $N_j$ is
the number of samples of node $j$.

As previously described
(see Subsection \ref{subsec:gb}),
when the loss function is different from the one used
when fitting the model to the matrix of residuals (as given by Eq. \eqref{eq:quality}).
a second optimization step has to be carried out as in
Eq. \eqref{eq:TerminalRegions}. In this second step, the $K$-dimensional vector of
outputs of the leaf nodes has to be updated as

\begin{equation}
    \gamma_{\{m,j,k\}} = \argmin_{\gamma} \sum_{\mathbf{x}_i \in \mathbb{R}_j} L(y_{ik}, F_{mk}(\mathbf{x}_i) + \gamma).
\end{equation}

For multi-class classification, if log-loss is used
\begin{equation}
    L(\mathbf{y}, \mathbf{F}(\mathbf{x})) = \sum_{k=1}^K y_k \ln \left (
    \frac{\exp(F_k(\mathbf{x}))}{\sum_{k=1}^{K} \exp(F_k(\mathbf{x}))}\right)\;,
\end{equation}
the Newton-Raphson update for tree terminal leaves would be
\begin{equation}
    \label{eq:TerminalRegions_clf}
    \gamma_{\{m, j, k\}} = \frac{\sum_{\mathbf{x}_i \in \mathbb{R}_{j}}^{}r_{\{i, k, m\}}}{\sum_{\mathbf{x}_i \in \mathbb{R}_{j}}^{} \big( (y_{\{i, k, m\}} - r_{\{i, k, m\}})  \times (1 -(y_{\{i, k, m\}} + r_{\{i, k, m\}})) \big)},
\end{equation}
using the first-order derivative of the
loss
function
(see Eq. \eqref{eq:TerminalRegions_clf})
near a given starting point to
generate a sequence of improved approximations to the root of the objective function.
For multi-output regression, if the square loss is used,
there is no need to update the terminal vector values since the trees are already optimizing the desired loss function.
The difference with a standard model would be that in the proposed algorithm all
multiple regression outputs are learned and predicted using one tree per iteration
and a single ensemble.
The pseudo-code for the proposed method is succinctly delineated in Algorithm \ref{alg:cgb}.

\begin{algorithm}
    \caption{Condensed Gradient Boosting}
    \label{alg:cgb}
    \begin{algorithmic}[1]
        \State \textbf{Input:}
        \begin{itemize}
            \item Training dataset $D = \{(\mathbf{x}_i, y_i)\}_{i=1}^N$.
            \item Loss function selection $L$.
            \item Number of boosting rounds $M$.
            \item Learning rate $\nu$.
        \end{itemize}

        \State \textbf{Output:}
        \begin{itemize}
            \item Trained multi-output Condensed-Gradient Boosting $\mathbf{{F}}_m(\mathbf{x})$.
        \end{itemize}

        \State Initialize $\mathbf{{F}}_0(\mathbf{x})$.
        \For{$t = 1$ to $M$}
        \State Compute the negative gradient.
        \State Fit the multi-output learner ($\mathbf{h}_m$) to the negative gradient Eq. \eqref{eq:pseudo_cgb}.
        \State Create splits that minimize the $L$ using Eq. \eqref{eq:quality}.
        \State Update terminal regions of $\mathbf{h}_m$ using Eq. \eqref{eq:TerminalRegions_clf}.
        \State Update the ensemble multi-output model using Eq. \eqref{eq:addmulti}.
        \EndFor
    \end{algorithmic}
\end{algorithm}

\subsection{Illustrative examples}

\begin{figure}[tb]
    \begin{tabular}{@{}c@{}c@{}}
        \subfloat[{\it C-GB}]{{\includegraphics[width=0.5\textwidth]{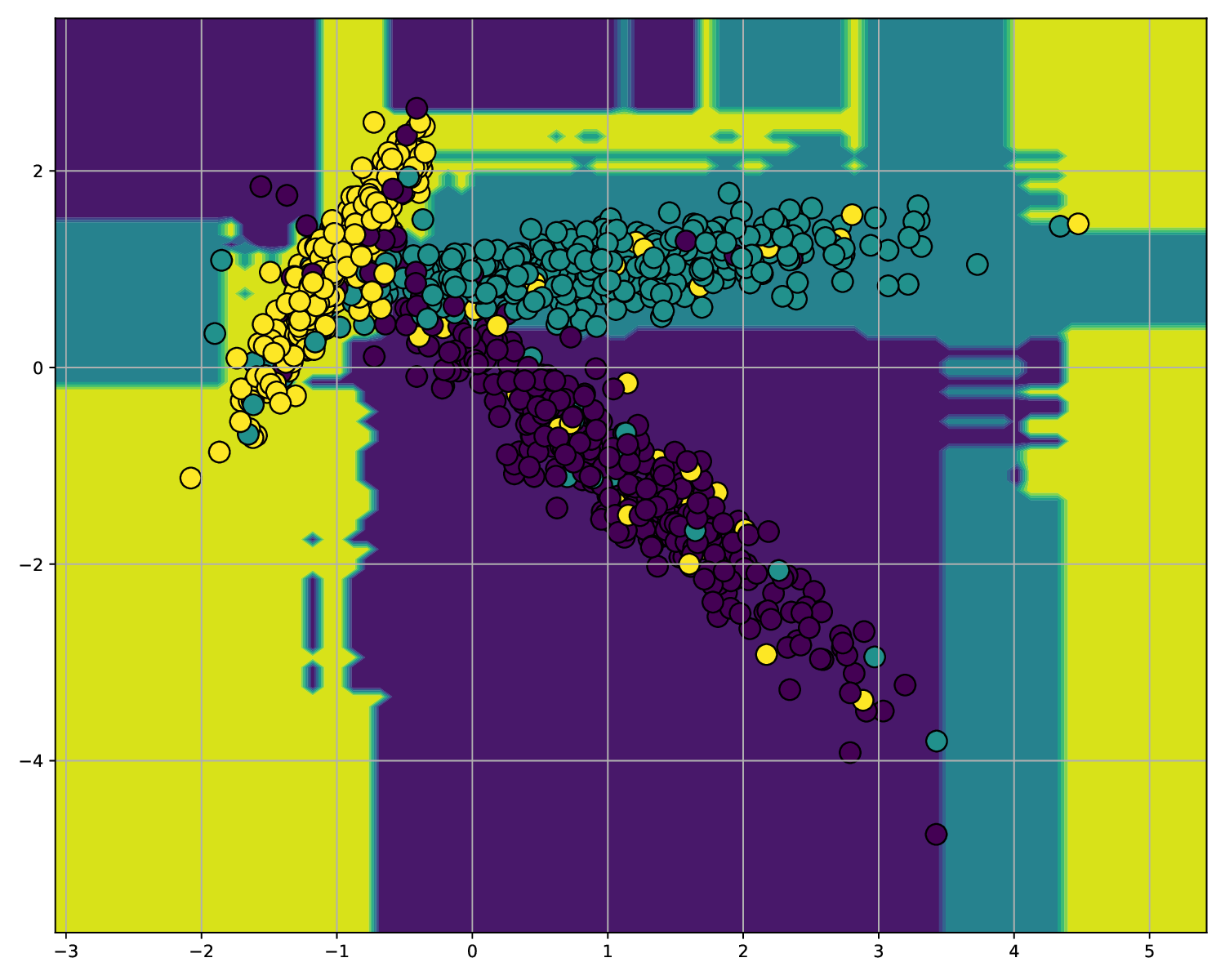}}} &
        \subfloat[{\it GB}]{{\includegraphics[width=0.5\textwidth]{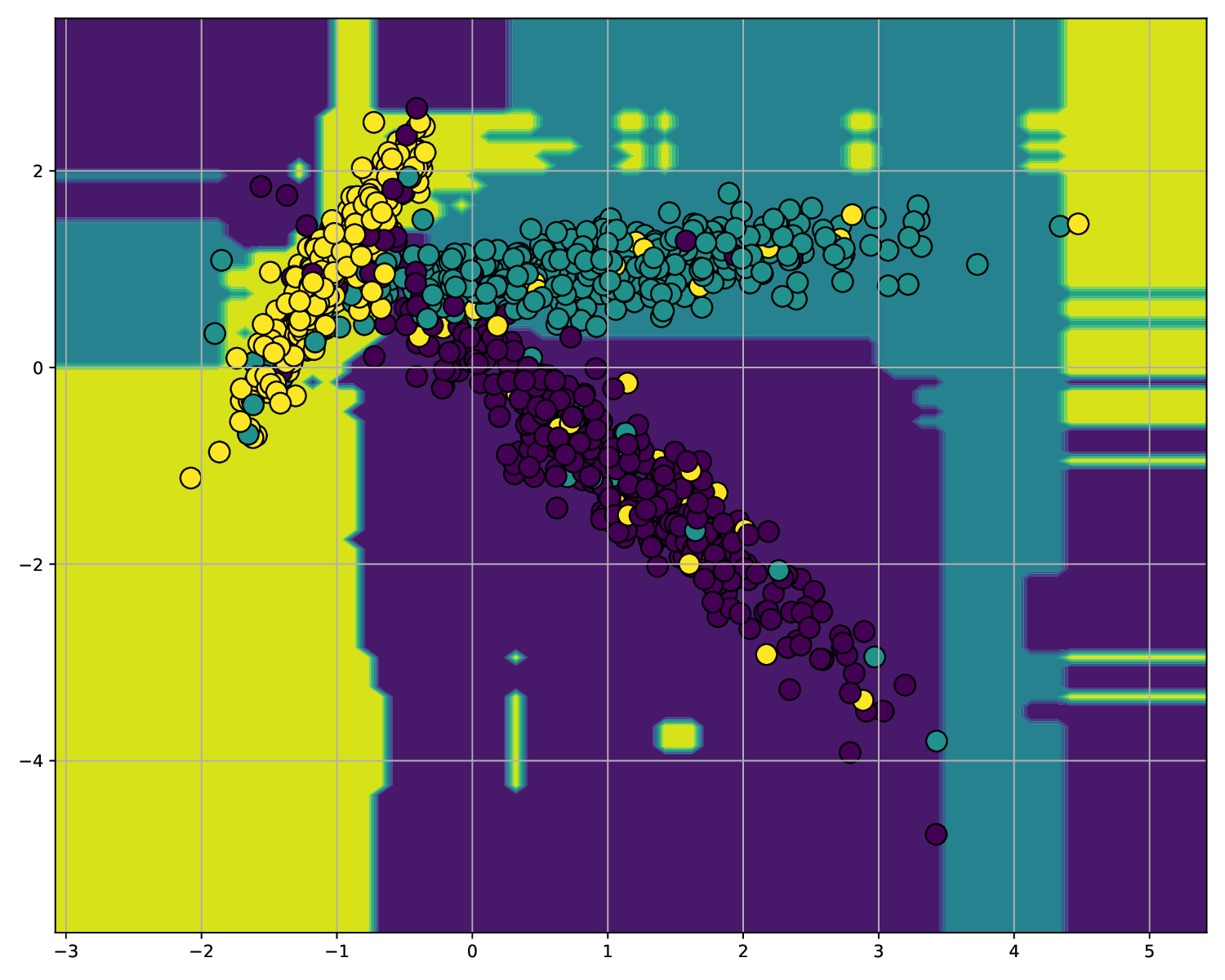}}}
    \end{tabular}
    \caption{Classification boundaries for the C-GB
        (subplot a) and GB models (subplot b)
    }
    \label{fig:DecisionBoundary}
\end{figure}

Below, we include various examples to demonstrate
the working mechanisms of the
proposed C-GB model and standard GB in multi-class
classification.
In the first experiment, we analyzed
the boundaries of GB and C-GB to compare
whether creating a single multi-output
regressor tree instead of one tree
per class could pose difficulties
in learning the concepts.
For this experiment, a synthetic
multi-class classification dataset with
three classe labels, $1200$ instances and two
input features is used.
The generated dataset is based on
the Madelon random data
experiment \cite{guyon2003design}.
The distribution of
the training data points is shown in
Figure \ref{fig:DecisionBoundary} with each
class in a different color.
The same hyperparameter configuration
is used for both methods (max depth=3,
subsample=0.75, learning rate=0.1, and
boosting iterations=100).
In Figure \ref{fig:DecisionBoundary},
the decision boundaries of the two methods are
shown using the same colors as the corresponding
training data points.
As evident from the subplots in
Figure \ref{fig:DecisionBoundary},
both C-GB (subplot a) and
GB (subplot b) exhibit remarkably similar performance
and adeptly adapt to the training instances.
This demonstrates the efficacy of
the proposed method in adapting to
the specific problem, despite training only
one tree per iteration.

\begin{figure}[tb]
    \centering
    \begin{tabular}{@{}c@{}c@{}}
        \subfloat[{\it C-GB}]{{\includegraphics[width=0.5\textwidth]{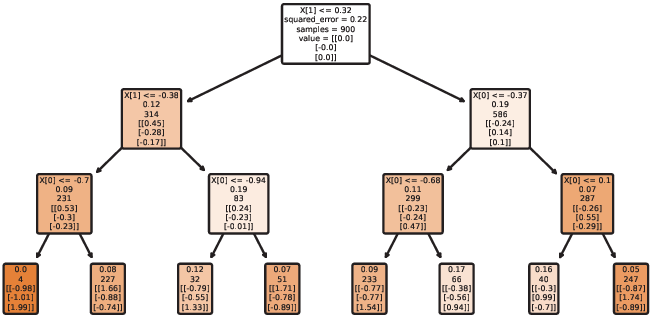}}}       &
        \subfloat[{\it GB class 0}]{{\includegraphics[width=0.5\textwidth]{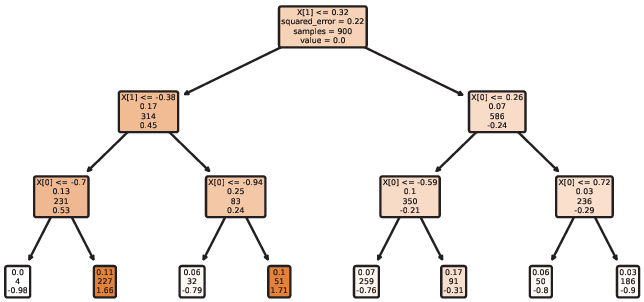}}}   \\
        \subfloat[{\it GB class 1}]{{\includegraphics[width=0.5\textwidth]{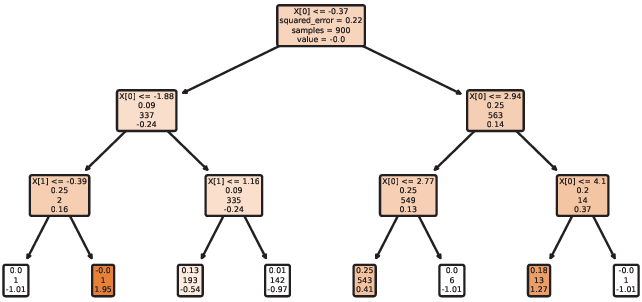}}} &
        \subfloat[{\it GB class 2}]{{\includegraphics[width=0.5\textwidth]{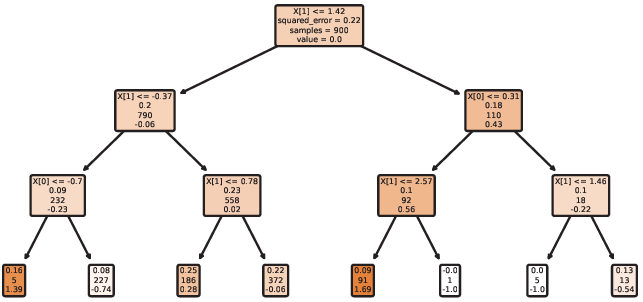}}}
    \end{tabular}
    \caption{Decision tree regressors for the C-GB (a),
        class zero of GB (b), first class of GB (c),
        and last class of GB (d)
        for a multi-class classification problem
        with three classes}
    \label{fig:DecisionTree}
\end{figure}

In Figure \ref{fig:DecisionTree},
the decision trees of the first iteration are shown as an
example of the classification task shown in
Figure \ref{fig:DecisionBoundary}.
The figure displays individual trees per
class for GB and a single tree for C-GB,
specifically for iteration one.
The tree nodes display various values:
split information for internal nodes,
MSE error of the node, the number of instances,
and the node output. In multi-output decision tree
(Figure \ref{fig:DecisionTree} subplot a),
the MSE represents the average error across all
outputs (as per Eq. \eqref{eq:quality}),
and the output value is a vector containing $K$ classes.
Figure \ref{fig:DecisionTree} (subplot a) illustrates how a
single tree can manage to
capture the information of the three classes.
The split of the root node and the
left branch is equivalent to the class $0$ tree
of GB (Figure \ref{fig:DecisionTree} subplot b), and
isolates the class $0$ (purple dots in
Figure \ref{fig:DecisionBoundary}) from the
rest. After the root node, the first two right
nodes of the C-GB tree have the
same splits as the first nodes of
the second class of GB tree
(Figure \ref{fig:DecisionTree} subplot c).
This illustrates how a
single tree is able to capture the information
of all classes in the problem in
a more compact way
(refer to Figure \ref{fig:DecisionTree} subplot a).
Also, using a single tree would produce more coherent
outputs as the same splits are used for all classes.

\begin{figure}[h!]
    \centering
    \begin{tabular}{@{}c@{}c@{}}
        \includegraphics[width=1\textwidth]{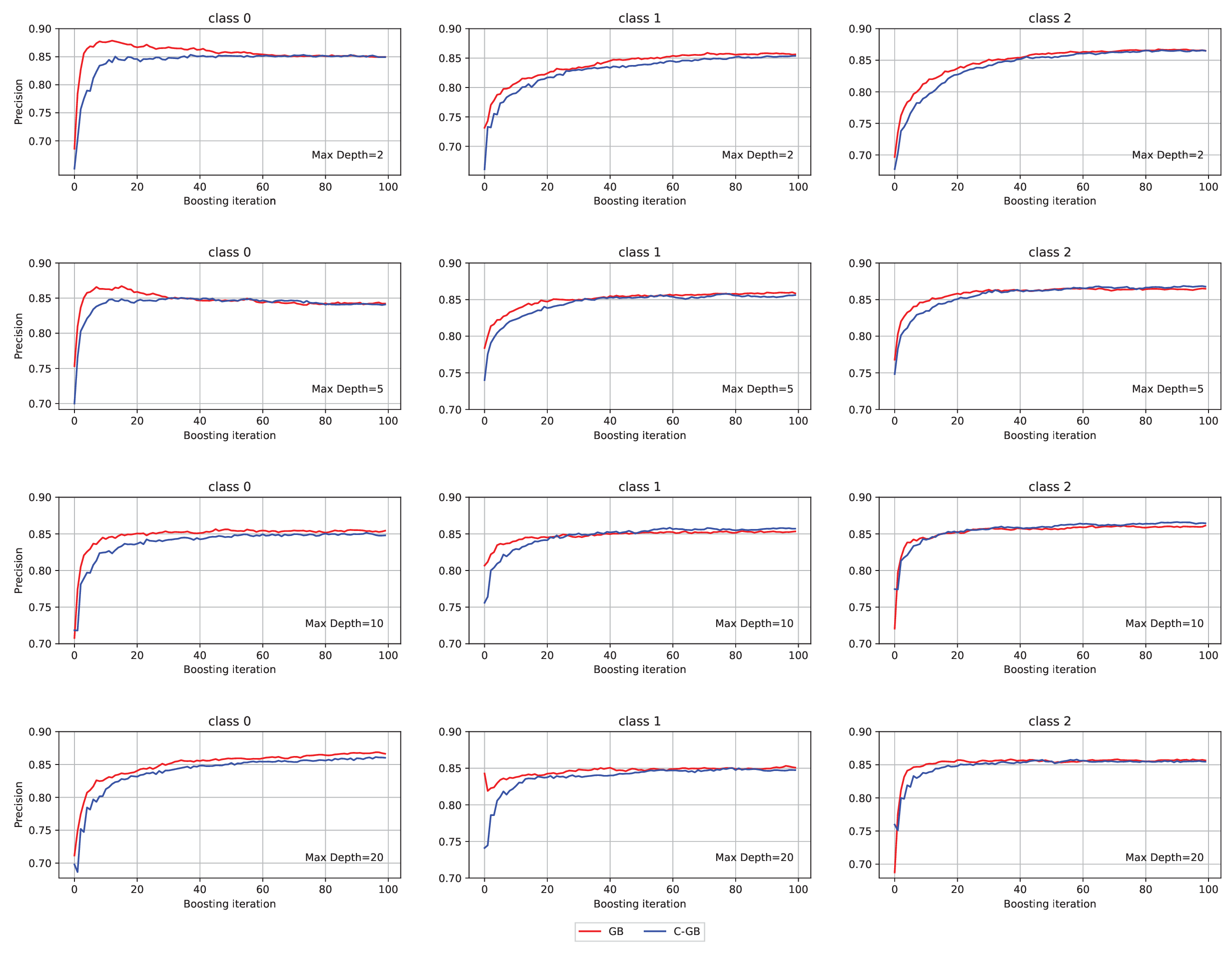}
    \end{tabular}
    \caption{The precision curves for C-GB
        (blue) and GB (red)
        are depicted for each class
        across boosting epochs, showcasing
        the impact of varying maximum depth
        values on the decision tree}
    \label{fig:Precision}
\end{figure}

To explore more the performance of
C-GB and GB, we analyzed their convergence
concerning the number of trees.
This assessment is conducted on the
    {\it Waveform} dataset, focusing
on the three class labels.
The experiments are repeated 10 times, and the
average values are reported.
The hyperparameters used are: learning rate is set to
0.1 and the subsampling rate to 0.75.
The convergence for each class is measured using
average precision (i.e. true positives
divided by detected positives) for each
class.
The results in average precision are shown
in Figure \ref{fig:Precision} for
various maximum depths
(2, 5, 10, and 20 tree depths, arranged in
ascending order from top to bottom)
for both C-GB
(blue curve) and GB (red curve).
From Figure \ref{fig:Precision}, it can be
observed that both methods achieve
equivalent precisions for all configurations.
Even for the smaller depths, the
final precision of both GB and C-GB are
almost identical although the
convergence for C-GB is a bit slower in this case.

\section{Experiments}
\label{sec:experiments}

In this section, we carry out an exhaustive comparison of the proposed C-GB with
respect to other similar algorithms such as TFBT \cite{ponomareva2017compact} and GBDT-MO \cite{zhang2020gbdt}.
In addition, the comparison is extended to single-output gradient boosting (GB).
The analyses are performed on 12 multi-class classification and three
multi-output regression tasks from diverse areas of application, each with
a different number of instances, classes, and attributes. The details of the
datasets are shown in Table \ref{table:Datasets}. We selected these datasets in
order to encompass significant variation in terms of different dimensions
(samples and features), as well as the number of classes and targets.

\begin{table*}[tb]
    \captionsetup{justification=raggedright,singlelinecheck=false}
    \caption{Dataset description}
    \resizebox{0.95\textwidth}{!}{%
        \begin{tabular}{lllll}
            \toprule
            Dataset                             & \# Samples & \# Features & \# Outputs & Training and test samples        \\
            \midrule
            {\bf Multi-class classification}                                                                               \\
            CIFAR-10 \cite{krizhevsky2010cifar} & 60,000     & 3,072       & 10         & 10,000 test, and 50,000 train    \\
            Cover Type \cite{UCI}               & 581,012    & 54          & 7          & 116,203 test, and 464,809 train  \\
            Digits \cite{UCI}                   & 1,797      & 64          & 10         & 360 test, and 1,437 train        \\
            iris \cite{UCI}                     & 150        & 4           & 3          & 30 test, and 120 train           \\
            Letter-26 \cite{UCI}                & 20,000     & 16          & 26         & 6,000, test and 14,000 train     \\
            MNIST \cite{deng2012mnist}          & 42,000     & 784         & 10         & 8,400 test, and 33,600 train     \\
            Poker Hand \cite{UCI}               & 1,025,010  & 15          & 10         & 25,010 test, and 1,000,000 train \\
            Sensit \cite{duarte2004vehicle}     & 98,528     & 100         & 3          & 19,705 test and 78,823 train     \\
            Vehicle \cite{UCI}                  & 846        & 18          & 4          & 170 test and 676 train           \\
            Vowel \cite{UCI}                    & 990        & 10          & 11         & 198 test and 792 train           \\
            Waveform \cite{UCI}                 & 5,000      & 21          & 3          & 1,000 test and 4000 train        \\
            Wine \cite{UCI}                     & 178        & 13          & 3          & 36 test and 142 train            \\
            \midrule
            {\bf Multi-output regression}                                                                                  \\
            ATP1d \cite{spyromitros2016multi}   & 337        & 411         & 6          & 68 test and 269 train            \\
            ATP7d \cite{spyromitros2016multi}   & 296        & 411         & 6          & 60 test and 236 train            \\
            Energy efficiency \cite{UCI}        & 768        & 8           & 2          & 154 test and 614 train           \\
            \bottomrule
        \end{tabular}\hspace*{-25pt}
    }
    \label{table:Datasets}
\end{table*}

The code of the proposed method (C-GB) is
implemented on top of the \\
\verb?scikit-learn? library and it
is available in the GitHub repository
\footnote{\href{https://github.com/GAA-UAM/C-GB}
    {github.com/GAA-UAM/C-GB}}.
For the single-output GB, we used the
implementation provided by
\verb?scikit-learn?,
(\verb?version 1.2 and older?)
\footnote{\href{https://github.com/scikit-learn/scikit-learn}
    {github.com/scikit-learn}}.
For GBDT-MO, we utilized the implementation
from the authors of the \verb?GBDTMO?
package \footnote{\href{https://github.com/zzd1992/GBDTMO}
    {github.com/zzd1992/GBDTMO}}, as referenced in \cite{zhang2020gbdt}.
The implementation of TFBT is
based on
TensorFlow \footnote{\href{https://git.kot.tools/nk2/syntaxnet_rus/-/tree/caae66a144f1237eb6b5c19fa00c317ca3bed09c/tensorflow/tensorflow/contrib/boosted_trees}{git.kot.tools/tensorflow}}.
Note, however, that in the last update of
TensorFlow, the TFBT package was
marked as deprecated
(\verb?version 2.8.0 and older?). Hence, we
used a previous version
of \verb?TensorFlow? (\verb?version 2.4.1?)
which was used in the corresponding
study by \cite{ponomareva2017compact}.
The libraries created by the author of
each method were used to faithfully
replicate the results.
Note that all these frameworks are
similar in the sense that they all provide a
python wrapper over a C/C++
implementation for the decision trees underneath,
which is the most time consuming operation
\cite{ke2017lightgbm}.
The libraries used by the tested
methods are all very well optimized for
speed, specially LightGBM and XGBoost
\cite{bentejac21comp}.

\subsection{Compared methods}
\label{subsec:Compared_methods}

The detailed definition of each of the algorithms included in the comparison and
their differences from our proposed method are described in this subsection.
Both the TFBT \cite{ponomareva2017compact} and
GBDT-MO \cite{zhang2020gbdt} methodologies are
developed on top of the XGBoost \cite{chen2016xgboost}
framework.
They both iteratively refine the optimization of a common objective function
(see Eq.\eqref{eq:obj_xgboost}). These methodologies are designed to manage
multi-output decision trees at each epoch ($\mathbf{h}_m$), with
$\mathbf{b}_j$ representing their vector size leaf ($j$) output,
given a designated loss function ($L$)
\begin{equation}
    \label{eq:obj_xgboost}
    \mathcal{L}_{m} =
    \sum_{i=1}^{N}\biggl [L(\mathbf{h}_m(\mathbf{x}_i),
    \mathbf{y}_i) + \mathbf{G}_i \mathbf{b}_{\{m, j\}} +
    \frac{1}{2} \mathbf{H}_i{\mathbf{b}_{\{m, j\}}}^2 \biggl ] +
    \Omega(\mathbf{b}_{\{m, j\}}),
\end{equation}
where, $\mathbf{G}_i$ and $\mathbf{H}_i$ denote the first and second-order derivative of the loss
function $L$ with respect to the decision tree outputs ($\mathbf{b}_{\{m,
        j\}}$), respectively, and $\Omega$ represents the ensemble regularizer, designed
to penalize the quantity of tree leaves as well as the $L2$ norms of their
respective weight vectors located at the leaf nodes.

The disparities between TFBT and GBDT-MO stem from their distinct
implementations in programming paradigms. TFBT extends XGBoost's foundational
architecture
\footnote{\href{https://github.com/dmlc/xgboost}
    {github.com/dmlc/xgboost}},
while GBDT-MO is constructed atop the LightGBM library
\footnote{\href{https://github.com/microsoft/LightGBM/}
    {github.com/microsoft/LightGBM}}.

\subsection{Experimental setup}

The validation procedure was carried out
using random splits with the
proportion of 20 percent for the test and the rest for the training sample, except for the
    {\it CIFAR-10}, {\it Poker Hand}, {\it MNIST} and {\it Sensit} for which the
partition determined by the datasets was used. In order to have stable results,
all experiments were repeated five times with different seed values except for
the largest datasets: {\it CIFAR-10}, {\it MNIST}, {\it Sensit}, {\it Poker
        Hand}, and {\it Cover Type} in which only one experiment was performed. The
reported generalization performance is computed as the average value of the
different realizations. The size of the training and test sets for each dataset
is indicated in the last column of Table \ref{table:Datasets}.
In an effort to achieve the highest possible accuracy performance for each
model, the hyperparameters of each model were tuned using the grid-search
method. This process involved conducting a grid search within the training phase
employing 2-fold cross-validation across all models and datasets. The
combination of hyperparameters yielding the best validation results was selected
to train the final model of each algorithm on the complete training dataset.
Table \ref{table:hyperparameters} displays the hyperparameter fixed values and
grids for each model. It should be noted that all models are configured with
boosting iterations.

\begin{table*}[tbp]
    \captionsetup{justification=raggedright,singlelinecheck=false}
    \caption{Hyperparameter grid reporting variable configurations
        and predetermined values across
        diverse models}
    \resizebox{1\textwidth}{!}{%
        \begin{tabular}{llllll}
            \toprule
            Model                 & Hyperparameter & Model                      & Hyperparameter                                                         \\
            \midrule
            \multirow{4}{*}{C-GB} & Tree depth     & [-1, 2, 5, 10, 20]         & \multirow{4}{*}{GB}       & Tree depth    & [-1, 2, 5, 10, 20]         \\
                                  & learning rate  & [0.025, 0.05, 0.1, 0.5, 1] & \multirow{4}{*}{}         & learning rate & [0.025, 0.05, 0.1, 0.5, 1] \\
                                  & max features   & [sqrt, None]               & \multirow{4}{*}{}         & max features  & [sqrt, None]               \\
                                  & subsample      & [0.75, 0.5, 1]             & \multirow{4}{*}{}         & subsample     & [0.75, 0.5, 1]             \\
            \midrule
            \multirow{3}{*}{TFBT} & Tree depth     & [2, 5, 10]                 & \multirow{3} {*}{GBDT-MO} & Tree depth    & [2, 5, 10, 20]             \\
                                  & learning rate  & [0.025, 0.05, 0.1, 0.5, 1] & \multirow{3}{*}{}         & learning rate & [0.025, 0.05, 0.1, 0.5, 1] \\
                                  & subsample      & [1]                        & \multirow{3}{*}{}         & subsample     & [0.75, 0.5, 1]             \\
            \bottomrule
        \end{tabular}
    }
    \label{table:hyperparameters}
\end{table*}

In addition to the generalization performance,
the training and prediction times
were also measured.
In training, only the time (in seconds) necessary to train the
models was logged without considering the time required to load the data in
memory. The prediction time was computed as the average time each model needs to
predict $10^5$ instances.
It is important to note that the training time depends on the hyperparameter
configuration.
To ensure consistency across models, we considered employing the same
set of hyperparameters for all models. These included subsample = $1$,
learning rate = $0.1$, max depth = $10$, and boosting iteration = $100$.
In addition, the complexity of the different models was conducted using big
$\mathcal{O}$ notation,
which is independent of the programming language of the implementation. This
measurement evaluates how the runtime and disk space consumption of an algorithm
increase relative to the input size and significant hyperparameters. It provides
an upper limit on the worst-case scenario for both runtime and space
requirements.
It is important to note that all experiments were conducted utilizing an
Intel(R) Xeon(R) CPU featuring six cores and operating at a base clock rate of
2.40GHz.

\subsection{Evaluation}

The generalization ability of the different algorithms
in the classification
tasks was assessed using the accuracy of the models.
For the multi-output regression problems,
the root mean square error ($RMSE$) was
used for each individual output,
\begin{align*}
    RMSE_k = \sqrt{\frac{\sum_{i=1}^{N} \; (y_{ik}-\hat{y}_{ik})^2}{N}},
\end{align*}
where $y_{ik}$ and $\hat{y}_{ik}$
are the target true and predicted values
respectively.
Furthermore, to have a unique performance
criterion for each model in the
multi-output regression tasks,
we considered the average coefficient of
determination, $aR^2$ over $K$ outputs
\begin{align*}
    aR^2 (y_{ik},\; \hat{y}_{ik}) = \sum_{k=1}^{K} \Big( 1- \sum_{i=1}^{N} \; \frac{(y_{ik}-\hat{y}_{ik})^2}{(y_{ik}-\bar{y}_{k})^2} \Big) / K,
\end{align*}
where $\bar{y}$ is the mean of the observed values $y_i$,
the highest possible value for $aR^2$ is $1.0$.
Negative values indicate poor performance,
worse than predicting the average of
the real target values for all test instances.
We did not consider the average $RMSE$
over all the targets as it would be an
average of values on different units,
making it difficult to interpret.

Finally, in order to compare the performance
across datasets and methods, the
statistical ranking methodology proposed by
Dems\v{a}r \cite{demvsar2006statistical} was applied.
This methodology computes the average
rank for each method across the tested
datasets. The difference in average rank
is considered significant using an Nemenyi
test. The result of this test can be plotted
using a Dems\v{a}r \cite{demvsar2006statistical}
as shown in Figure \ref{fig:demsar}. The
ranking of the methods in each dataset
was done using the average accuracy in
the classification tasks.
For the regression problems, the methods were
ranked by target using the $RMSE$ metric.

\subsection{Results}
\label{subsec:results}

The average results for the different experiments
are shown in
Tables \ref{table:ClfPerformance},
\ref{table:RMSEPerformance},
and \ref{table:R2Performance}.
In Table \ref{table:ClfPerformance}, the average
accuracy for the multi-class tasks is shown.
Tables \ref{table:RMSEPerformance} and
\ref{table:R2Performance} show the
average $RMSE$ and average
${R^2}$-score for the multi-output regression
problems.
The tables show the comparative e
valuation of the following methods:
the proposed C-GB model, Standard single-output Gradient Boosting (GB), TFBT
(only for classification), and GBDT-MO.
The best value for each dataset or target is highlighted in the tables with a
light blue shading.

As shown in Table \ref{table:ClfPerformance}, GB generally exhibits an accuracy
superior to that of the other models, achieving the highest accuracy in eight
out of 12 datasets.
The C-GB shows a rather good performance with four best performances. GBDT-MO
obtained the best performance in one dataset, and TFBT did not achieve the best
performance in any of the datasets.
The disparity in performance between GB and its extension, C-GB, is
generally marginal across most datasets. For example, in the {\it Waveform}
dataset, the accuracy difference between these models is approximately $0.12$,
and in {\it Cover Type}, it is around $0.02$. The most significant
performance margin in favor of GB occurs in the {\it Poker Hand} dataset, with a
notable difference of $1.94$ points, coinciding with the dataset where GBDT-MO
achieves its best result. C-GB demonstrates its highest performance in {\it
        Wine} and {\it Vowel}, surpassing GB by $1.11$ and $0.4$ points, respectively.

\begin{table*}[tb]
    \captionsetup{justification=raggedright,singlelinecheck=false}
    \caption{Average generalization accuracy for C-GB
        , GB, TFBT,
        and GBDT-MO. The best results are highlighted in a light blue hue}
    \begin{tabularx}{1\textwidth}{*{5}X}
        \toprule
        Dataset    & C-GB                         & GB                           & TFBT  & GBDT-MO                      \\
        \midrule
        CIFAR-10   & 51.31                        & \cellcolor{lightblue}{51.98} & 36.15 & 48.67                        \\
        Cover Type & 97.23                        & \cellcolor{lightblue}{97.25} & 73.94 & 84.60                        \\
        Digits     & 97.17                        & \cellcolor{lightblue}{97.39} & 95.94 & 96.39                        \\
        Iris       & 94.00                        & \cellcolor{lightblue}{95.33} & 93.33 & 92.67                        \\
        Letter-26  & \cellcolor{lightblue}{96.80} & 96.57                        & 76.44 & 94.74                        \\
        MNIST      & 97.48                        & \cellcolor{lightblue}{97.63} & 80.48 & 96.71                        \\
        Poker Hand & 74.87                        & 76.81                        & 57.63 & \cellcolor{lightblue}{78.15} \\
        Sensit     & 84.83                        & \cellcolor{lightblue}{85.07} & 75.47 & 82.95                        \\
        Vehicle    & \cellcolor{lightblue}{75.06} & \cellcolor{lightblue}{75.06} & 64.24 & 72.12                        \\
        Vowel      & \cellcolor{lightblue}{96.16} & 95.76                        & 70.91 & 83.54                        \\
        Waveform   & 85.12                        & \cellcolor{lightblue}{85.24} & 80.00 & 84.40                        \\
        Wine       & \cellcolor{lightblue}{98.89} & 97.78                        & 92.78 & 86.67                        \\
        \bottomrule
    \end{tabularx}

    \label{table:ClfPerformance}
\end{table*}

The results of the multi-output regression experiments are shown in
Tables. \ref{table:RMSEPerformance} and \ref{table:R2Performance}. In this
experiment, as GB does not support multi-output regression, we trained one
different ensemble for each output separately. Results are not reported for TFBT
as this method cannot handle regression tasks.

From Table. \ref{table:RMSEPerformance}, it can be observed that the performance
of C-GB and GB on each target are and superior to that of GBDT-MO.
Specifically, GB achieves the optimal performance in nine targets,
whereas C-GB demonstrates superiority in five. Conversely, GBDT-MO fails to
achieve comparable performance to the other two methodologies. A similar trend
is observed in the average $R^2$ score shown in Table \ref{table:R2Performance}.
GB attains the highest score across all datasets, followed by C-GB.

\begin{figure}[tb]
    \includegraphics[width=\textwidth]{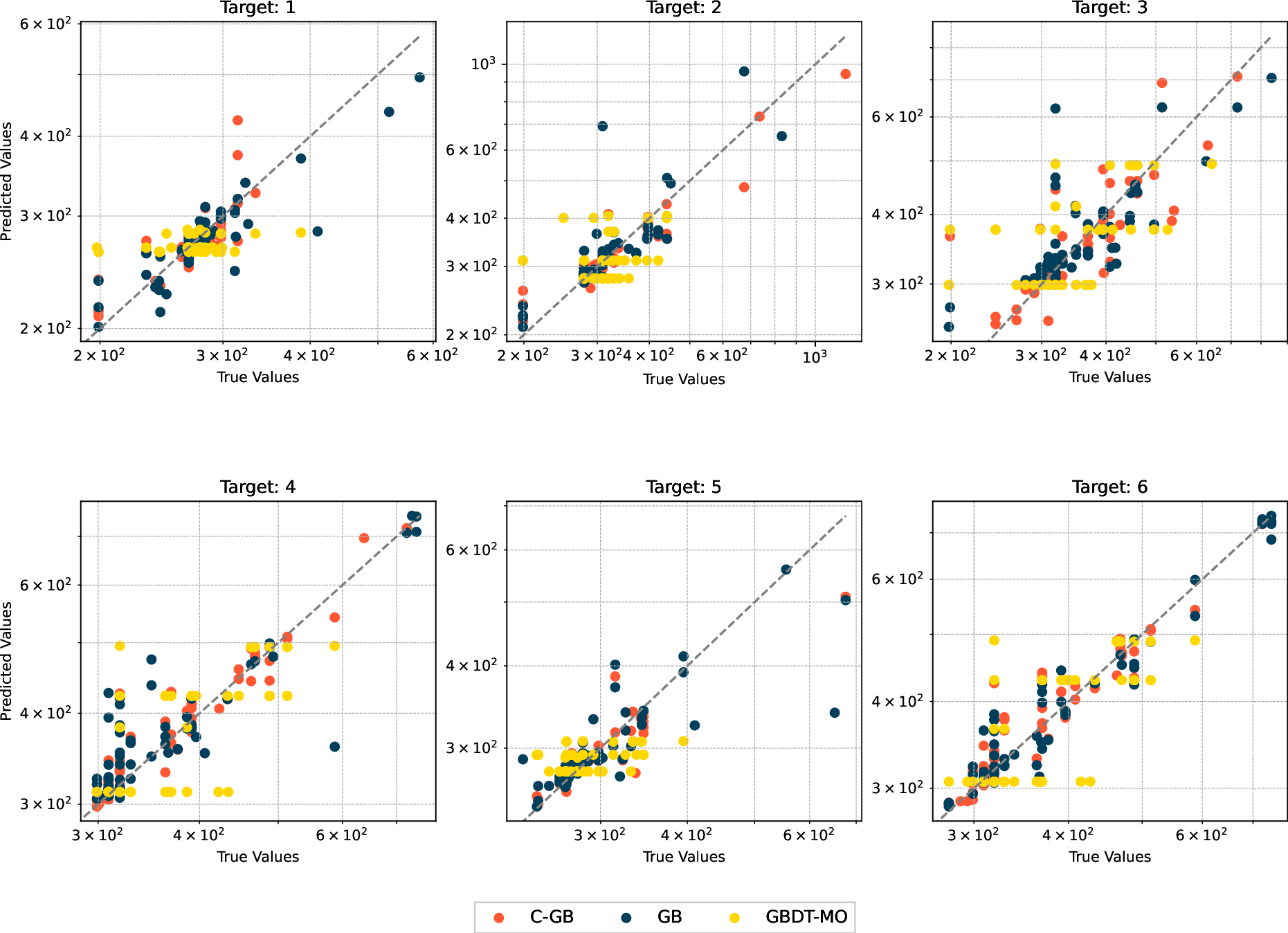}
    \caption{Scatter Diagram of Prediction:
        For six outputs of the ATP7D
        and three models: C-GB (red dots),
        GB (dark blue dots),
        and GBDT-MO (yellow dots)}
    \label{fig:scatter}
\end{figure}

In addition, the average performance of GBDT-MO in the {\it ATP1D} dataset is
clearly lower than that of the other two methods. To explore in the dynamics od
this dataset, a scatter plot illustrating the correspondence between the
predicted and actual values for each target is shown in Figure \ref{fig:scatter}.
The poor performance of GBDT-MO in this dataset can be attributed to its
tendency to generate similar predictions for most instances.

\begin{table*}[tb]
    \captionsetup{justification=raggedright,singlelinecheck=false}
    \caption{Average generalization RMSE for C-GB,
        GB, and GBDT-MO.
        The best results are highlighted in a light blue hue}
    \resizebox{1.0\textwidth}{!}{%
        \begin{tabular}{l*{6}c|*{6}c|cc}
            \toprule
            \multicolumn{1}{l}{Dataset} & \multicolumn{6}{c}{ATP1d}    & \multicolumn{6}{c}{ATP7d}     & \multicolumn{2}{c}{Energy}                                                                                                                                                                                                                                                                                                                                                      \\

            \multicolumn{1}{l}{Target}  & 0                            & 1                             & 2                            & 3                            & 4                            & 5                            & 0                            & 1                            & 2                            & 3                            & 4                            & 5                            & 0                           & 1                           \\
            \midrule
            C-GB                        & 51.98                        & \cellcolor{lightblue}{107.79} & 79.58                        & 51.74                        & 61.99                        & 59.48                        & \cellcolor{lightblue}{25.39} & 68.86                        & \cellcolor{lightblue}{55.72} & 39.03                        & \cellcolor{lightblue}{33.34} & 41.22                        & \cellcolor{lightblue}{0.37} & 0.94                        \\
            GB                          & \cellcolor{lightblue}{48.38} & 113.19                        & \cellcolor{lightblue}{74.55} & \cellcolor{lightblue}{32.26} & \cellcolor{lightblue}{51.27} & \cellcolor{lightblue}{34.24} & 28.09                        & \cellcolor{lightblue}{64.97} & 61.42                        & \cellcolor{lightblue}{37.78} & 34.98                        & \cellcolor{lightblue}{35.41} & 0.49                        & \cellcolor{lightblue}{0.83} \\
            GBDT-MO                     & 77.04                        & 171.22                        & 136.08                       & 129.41                       & 90.34                        & 134.34                       & 97.03                        & 190.15                       & 171.74                       & 179.38                       & 111.71                       & 184.78                       & 1.21                        & 2.10                        \\
            \bottomrule
        \end{tabular}
    }
    \label{table:RMSEPerformance}
\end{table*}

\begin{table*}[h!]
    \captionsetup{justification=raggedright,singlelinecheck=false}
    \caption{Average generalization coefficient of
        determination C-GB, GB-SO, GBDT-MO.
        The best results are highlighted in a light blue hue}
    \begin{tabularx}{0.99\textwidth}{*{4}X}
        \toprule
        {Dataset} & C-GB   & GB                            & GBDT-MO \\
        \midrule
        ATP1d     & 0.8210 & \cellcolor{lightblue}{0.8381} & 0.4567  \\
        ATP7d     & 0.6952 & \cellcolor{lightblue}{0.7203} & -1.4765 \\
        Energy    & 0.9944 & \cellcolor{lightblue}{0.9982} & 0.9606  \\
        \bottomrule
    \end{tabularx}
    \label{table:R2Performance}
\end{table*}

\begin{figure}[tb]
    \centering
    \begin{tabular}{@{}c@{}c@{}}
        \subfloat[Multi-class
        classification]{{\includegraphics[width=0.5\linewidth]{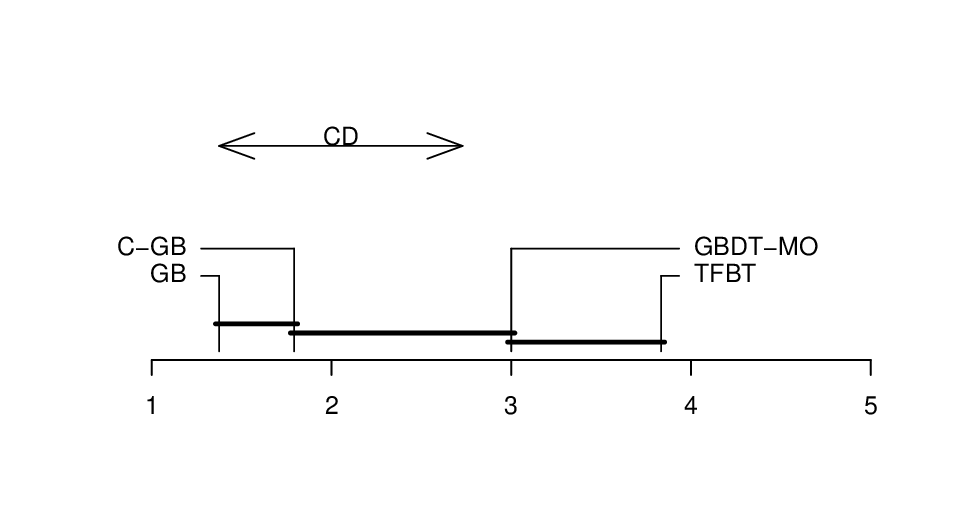}}} &
        \subfloat[Regression-Target wise]{{\includegraphics[width=0.5\linewidth]{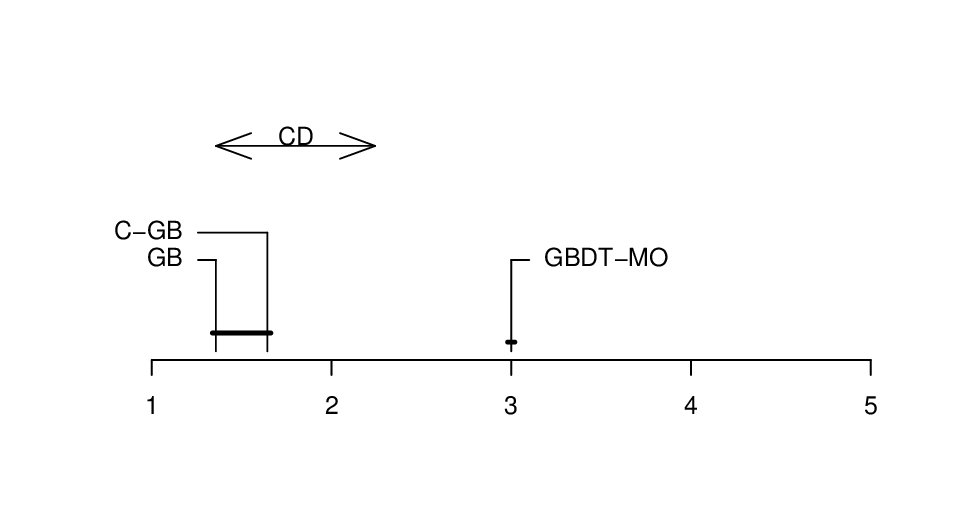}}}
    \end{tabular}
    \caption{Comparison of different
        gradient boosting models (higher rank is better)
        using the Nemenyi Test, ($p = 0.05$)
    }
    \label{fig:demsar}
\end{figure}

Figure \ref{fig:demsar} shows the overall comparison of the tested methods
using a Dems\v{a}r diagram. The plots display the average rank of the methods for
the analyzed datasets. The difference between the two methods is statistically
significant if they are not connected with a horizontal solid line. The critical
distances (CD) are 1.64 and 0.88 for 12 multi-class classification tasks and 14
regression targets, respectively with $p\text{-value} = 0.05$.

Similar conclusions can be drawn from these diagrams. For the multi-class
classification problems (Figure \ref{fig:demsar} subplot a), the worse
performance is produced by TFBT followed by GBDT-MO. The best average rank is
achieved by GB followed by the proposed C-GB model. The performance of C-GB in
multi-class classification is statistically significantly better than that of
TFBT. As for the regression tasks (Figure \ref{fig:demsar} subplot b), we ranked,
the models over 14 regression targets for three models. The best rank was
achieved by the GB and C-GB with statistical significant difference in relation
to GBDT-MO.
It is noteworthy that GBDT-MO differs statistically
from both C-GB and GB models.

One critical aspect of these methods is their computational complexity. To
assess this aspect, two different analyses are employed. First, the training and
inference (test) times of the studied models,
are measured for all datasets. For this analysis the implementation of the
authors was used for each method. Second, and in order to have a complexity
analysis independent of their implementation, we carried out a second analysis
based on big $\mathcal{O}$ notation. With this analysis the time and space
complexities of these models was explored.

The time performances are outlined in
Table \ref{table:TrainingTimes}. The
results of this table are generally more favorable to two methods based on
multi-output trees: C-GB and GBDT-MO. C-GB
trained faster in eight datasets, and the GBDT-MO trained faster in seven. In
addition, the training times are generally favorable to C-GB in the largest
datasets: for instance in {\it CIFAR-10} and {\it MNIST}, GBDT-MO is
respectively $2528.42$ and $861.12$ seconds slower than the C-GB. The training
times for GB are clearly slower than those of C-GB. Regarding the test time
required to evaluate $10^5$ instances, the GBDT-MO model was the fastest in nine
datasets and C-GB in four. The results of GB, in this regard, are somewhat more
competitive in small datasets, even though one tree per class and iteration has
to be evaluated. The reason for this is that in general, the size of the
decision trees generated by GB is smaller.

\begin{table*}[tb]
    \captionsetup{justification=raggedright,singlelinecheck=false}
    \caption{Training and prediction time
        comparison (in seconds) for
        C-GB, GB, TFBT, and GBDT-MO.
        The best results are highlighted in light blue hue}
    \resizebox{0.95\textwidth}{!}{%
        \begin{tabular}{lrr|rr|rr|rr}
            \toprule
            \multicolumn{1}{l}{Models} & \multicolumn{2}{c}{C-GB}      & \multicolumn{2}{c}{GB}                 & \multicolumn{2}{c}{TFBT} & \multicolumn{2}{c}{GBDT-MO}                                                                                                        \\
            \multicolumn{1}{l}{Datset} & Training                      & Prediction                             & Training                 & Prediction                  & Training  & Prediction                  & Training                     & Prediction                  \\
            \midrule
            \textbf{Multi-class classification}
            CIFAR-10                   & \cellcolor{lightblue}{332.69} & 48.29                                  & 2274.58                  & 42.60                       & 138265.17 & 643.44                      & 2861.12                      & \cellcolor{lightblue}{8.42} \\
            Cover Type                 & \cellcolor{lightblue}{249.86} & \cellcolor{lightblue}{lightblue}{1.55} & 1092.56                  & 4.78                        & 67268.82  & 7.51                        & 345.41                       & 3.46                        \\
            Digits                     & \cellcolor{lightblue}{2.72}   & 5.51                                   & 14.07                    & 19.16                       & 1687.25   & 262.56                      & 4.60                         & \cellcolor{lightblue}{2.88} \\
            iris                       & 0.18                          & 32.38                                  & 0.35                     & 19.21                       & 26.11     & 1560.24                     & \cellcolor{lightblue}{0.04}  & 1.22                        \\
            Letter-26                  & 15.30                         & \cellcolor{lightblue}{2.16}            & 157.23                   & 23.67                       & 2399.45   & 16.75                       & \cellcolor{lightblue}{13.78} & 3.72                        \\
            MNIST                      & \cellcolor{lightblue}{90.89}  & 15.93                                  & 687.03                   & 33.00                       & 34392.61  & 173.45                      & 952.02                       & \cellcolor{lightblue}{5.81} \\
            Poker Hand                 & \cellcolor{lightblue}{12.29}  & 2.33                                   & 71.59                    & 9.31                        & 334.97    & \cellcolor{lightblue}{1.59} & 35.02                        & 2.39                        \\
            Sensit                     & 132.23                        & \cellcolor{lightblue}{2.17}            & 325.21                   & 2.64                        & 2020.74   & 22.03                       & \cellcolor{lightblue}{73.17} & 3.63                        \\
            Vehicle                    & 1.21                          & 7.95                                   & 3.29                     & 10.27                       & 133.00    & 325.85                      & \cellcolor{lightblue}{0.90}  & \cellcolor{lightblue}{1.81} \\
            Vowel                      & 2.06                          & 7.94                                   & 10.98                    & 25.26                       & 390.70    & 277.87                      & \cellcolor{lightblue}{1.13}  & \cellcolor{lightblue}{2.22} \\
            Waveform                   & 3.68                          & \cellcolor{lightblue}{2.31}            & 7.49                     & 3.14                        & 159.74    & 64.57                       & \cellcolor{lightblue}{2.72}  & 3.55                        \\
            Wine                       & 0.21                          & 32.19                                  & 0.39                     & 16.99                       & 73.39     & 1281.90                     & \cellcolor{lightblue}{0.10}  & \cellcolor{lightblue}{0.99} \\
            \midrule
            \textbf{Multi-output regression}
            ATP1d                      & \cellcolor{lightblue}{0.34}   & 18.22                                  & 1.21                     & 21.23                       &           &                             & 8.17                         & \cellcolor{lightblue}{1.19} \\
            ATP7d                      & \cellcolor{lightblue}{0.30}   & 19.82                                  & 1.07                     & 21.46                       &           &                             & 7.22                         & \cellcolor{lightblue}{1.12} \\
            Energy                     & \cellcolor{lightblue}{0.14}   & 6.97                                   & 0.16                     & 4.16                        &           &                             & 0.20                         & \cellcolor{lightblue}{1.28} \\
            \bottomrule
        \end{tabular}\hspace*{-25pt}
    }
    \label{table:TrainingTimes}
\end{table*}

In relation to time complexity of the algorithms, we focus on the time required
to fit the decision tree structure, where the most complex operations occur
\cite{ke2017lightgbm}.
For that, we
revisit the key parameters in our study: $N$ represents the total number of
samples, $P$ denotes the number of input features and $\lambda$ the depth of
the trees.
The time complexity for building a decision tree can be expressed
in terms of the complexity of the split finding algorithm and the number of
splits related to the depth. In this study, the various models differ
in terms of the algorithms used to find the best splits.
For CART-based decision tree splits, which are the trees used in GB and C-GB,
the split algorithm consists of ordering the values of an attribute
($N \log_2(N)$) and then loop over it ($N$) to compute the gain of each possible
split (this operation can be performed independently of the number of classes
although there is a one time cost $K$ of initialization).
This is repeated for all attributes ($P$), so the complexity is
$\mathcal{O}(P \times (N \log_2(N) + N + K)) =
    \mathcal{O}(P \times N \log_2(N))$, assuming $K<<N$.
Considering the data is divided in half after each split, we have that at the next
depth $2 \times P \times N/2 \log_2(N/2)$ operations need to be done, which
resolves to $\mathcal{O}(P \times N \times \log_2(N))$. Hence, the total cost for
building a tree is $\mathcal{O}(P \times N \times \log_2(N) \times \lambda)$.
On the other hand, GBDT-MO utilizes the LightGBM implementation
and TFBT is based on XGBoost.
Both LightGBM and XGBoost exhibit
identical time complexity in split finding \cite{zhang2020gbdt}
characterized by $\mathcal{O}(\beta \times P \times K)$ where $\beta$
is the number of bins employed within the split-finding histogram
methodology. Further details and substantiation can be found in
\cite{zhang2020gbdt}. This splitting cost is the same for every level of the
tree. Consequently, the cost for building a tree is $\mathcal{O}(\beta \times
    P \times K \times \lambda^2)$. This algorithm also has a one time cost of
$P\times N$ that is needed to compute the bins.
With that said and recalling the number of boosting
epochs ($M$) and classes ($K$), we can express the total time complexity of each
analyzed model
as shown in Table \ref{table:TimeComplexity}.

\begin{table}[tb]
    \captionsetup{justification=raggedright,singlelinecheck=false}
    \caption{Time complexity}
    \centering
    \begin{tabular}{ll}
        \toprule
        {Model} & {Time complexity}                                                            \\
        \midrule
        C-GB    & $\mathcal{O}(M \times P \times N \times \log_2(N) \times \lambda)$           \\
        GB      & $\mathcal{O}(K \times M \times P \times N \times \log_2(N) \times \lambda)$  \\
        TFBT    & $\mathcal{O}(M \times \beta \times P \times K \times \lambda^2 + P\times N)$ \\
        GBDT-MO & $\mathcal{O}(M \times \beta \times P \times K \times \lambda^2 + P\times N)$ \\
        \bottomrule
    \end{tabular}
    \label{table:TimeComplexity}
\end{table}

As shown in Table~\ref{table:TimeComplexity},
the models based on bins ($<<N$) have a better complexity with respect to GB and
C-GB, specially for small depths ($\lambda$).
In any case, depending on the selected hyperparameters
selected for each model the actual training times could differ in practice.

To assess the space complexity of the studied algorithms, we outline three
fundamental allocations necessary for training the models as follows:
\begin{itemize}
    \item Storage of decision trees:
          $\mathcal{O}(M \times |h(\mathbf{x})|)$, where
          $\mathcal{O}(|h(\mathbf{x})|)$ is the tree size that is given by
          the number of nodes times the size of each node. The number of nodes
          is $\mathcal{O}(\lambda^2))$. The node size includes
          the split threshold, attribute index, pointers to the left and right
          nodes, and, for leave nodes, the output values, which is a $K$-size
          vector. In consequence, the size of a tree is
          $\mathcal{O}(h(\mathbf{x}))=\mathcal{O}(\lambda^2 \times (K +
              \#node\_params))$
    \item Gradient information:
          $\mathcal{O}(N \times K \bigl)$.
    \item A space reserve dedicated to housing histograms (specifically in the
          case of GBDT-MO and TFBT algorithms):
          $\mathcal{O}(\beta \times K)$.
\end{itemize}
This gives a space complexity of the analyzed models shown in
Table~\ref{table:SpaceComplexity}. The space complexity of GB is the largest of
the models, especially considering that the size required for the number of
parameters of a tree node is usually larger than the number of classes.

In fact, one of the main benefits of multi-output
models is their fast prediction speed, as illustrated in Table
\ref{table:TrainingTimes}.

\begin{table}[tb]
    \captionsetup{singlelinecheck=false, justification=raggedright}
    \caption{Sapce complexity}
    \centering
    \begin{tabular}{ll}
        \toprule
        {Model} & {Space complexity}                                                                              \\
        \midrule
        C-GB    & $\mathcal{O}(M \times (\lambda^2 \times (K + \#node\_params)) + N \times P \times K) $          \\
        GB      & $\mathcal{O}(K \times M \times (\lambda^2 \times (2 + \#node\_params)) + N \times P \times K) $ \\
        TFBT    & $\mathcal{O}(M \times (\lambda^2 \times (K + \#node\_params)) + N \times P
        \times K + \beta \times K) $                                                                              \\
        GBDT-MO & $\mathcal{O}(M \times (\lambda^2 \times (K + \#node\_params)) + N \times P
        \times K + \beta \times K) $                                                                              \\
        \bottomrule
    \end{tabular}
    \label{table:SpaceComplexity}
\end{table}

\section{Conclusion}
\label{sec:conclusion}
In this paper, Condensed Gradient Boosting (C-GB) is presented. C-GB is a
variant of Gradient Boosting (GB) for multi-class classification and
multi-output regression problems. In multi-class problems (i.e. tasks with more
than two classes), GB has to train one regressor model per class and iteration.
In this work, we proposed the use of decision trees with vector-valued leaves in
such a way that only one decision tree per iteration has to be trained. We have
shown that these multi-output regression trees are able to capture the
information of all classes at once. In addition, the use of these trees allowed
us to adapt the model to multi-output regression. The proposed method is
implemented as an open-source package for further investigations.
An extensive experimental comparison has been carried out. The proposed model
has been compared with standard GB.
Moreover, this comparison
was extended to include two other similar multi-output models employing distinct
optimization strategies: TFBT utilizing XGBoost and GBDT-MO employing LightGBM.
The experiments showed that the proposed approach achieved a level of
generalization performance comparable to that of the standard GB method, while
decreasing the utilization of computational resources. The ensembles
produced by C-GB generally outpace standard GB in training and prediction speed.
Moreover, the proposed model exhibits superior generalization performance when
compared to TFBT and GBDT-MO in the tested dataset. In computational terms, TFBT
emerges as the slowest method, while GBDT-MO and C-GB stand out as the fastest
among all tested techniques.

The proposed model's performance has significant implications for real-world
applications, particularly in terms of computational efficiency. Future research
should prioritize assessing its compatibility with deep learning architectures
as well as exploring techniques to handle imbalanced classes and
high-dimensional feature spaces, common challenges in multi-class problems.

\section*{Acknowledgment}
The authors acknowledge financial support from project PID2022-139856NB-I00
funded by MCIN/ AEI / 10.13039/501100011033 / FEDER, UE and project
PID2019-106827GB-I00 / AEI / 10.13039/501100011033 and from the Autonomous
Community of Madrid (ELLIS Unit Madrid).
We also extend our appreciation to the Centro
de Computación Científica CCC-UAM
for providing computational resources.

\bibliographystyle{unsrt}  
\bibliography{ref}

\end{document}